\documentclass{article}

\usepackage{PRIMEarxiv}

\usepackage[utf8]{inputenc} 
\usepackage[T1]{fontenc}    
\usepackage{hyperref}       
\usepackage{url}            
\usepackage{booktabs}       
\usepackage{amsfonts}       
\usepackage{nicefrac}       
\usepackage{microtype}      
\usepackage{lipsum}
\usepackage{fancyhdr}       
\usepackage{graphicx}       
\graphicspath{{media/}}     

\title{Jalisco's multiclass land cover analysis and classification using a novel lightweight convnet with real-world multispectral and relief data
}

\author{
  Alexander Quevedo, Abraham Sánchez, Raul Nancláres, Diana P. Montoya, \\
  \textbf{Juan Pacho, Jorge Martínez, and E. Ulises Moya-Sanchez*} \\
  Dirección General de Inteligencia Gubernamental \\
  Coordinación general de innovación Gubernamental del Gobierno del estado Jalisco \\
  Guadalajara, Jalisco México\\
  \texttt{*eduardo.moya@jalisco.gob.mx} \\
  
}

\begin{document}
\maketitle

\begin{abstract}
The understanding of global climate change, agriculture resilience, and deforestation control rely on the timely observations of the Land Use and Land Cover Change (LULCC).  Recently,  some deep learning (DL) methods have been adapted to make an automatic classification of  Land Cover (LC) for global and homogeneous data. However,  most of these DL models can not apply effectively to real-world data. i.e. a large number of classes, multi-seasonal data,  diverse climate regions, high imbalance label dataset, and low-spatial resolution. In this work, we present our novel lightweight (only 89k parameters) Convolution Neural Network (ConvNet)  to make LC classification and analysis to handle these problems for the Jalisco region. In contrast to the global approaches, the regional data provide the context-specificity that is required for policymakers to plan the land use and management, conservation areas, or ecosystem services.  In this work, we combine three real-world open data sources to obtain 13 channels. Our embedded analysis anticipates the limited performance in some classes and gives us the opportunity to group the most similar, as a result, the test accuracy performance increase from 73 \% to 83 \%. We hope that this research helps other regional groups with limited data sources or computational resources to attain the United Nations Sustainable Development Goal (SDG) concerning “Life on Land”.
\end{abstract}


\section{Introduction}

Terrestrial vegetation is a critical component of global biogeochemical cycles and provides important ecosystem services to support human life~\cite{pastur2018ecosystem}. Given its importance, it is essential to know the spatial-temporal variations of vegetation \cite{gao2021monitoring}. These variations are due to several determining factors such as global climate variability, climate gradients, and anthropogenic factors such as Land Use and Land Cover Change (LULCC).  The diversity in climatic conditions and vegetation types pose different obstacles to monitoring and classifying land cover using remote sensing. Mexico is considered one of the mega-diverse countries on the planet due to its location in a transition zone between Nearctic and Neotropic regions making it more difficult for land use classification and monitoring.

The anthropogenic factors,  could be a trigger for deforestation and forest degradation~\cite{franklin2016global} and have a severe impact on the global carbon cycle, soil erosion, hydrological cycles, and in general, affect on the ecosystem services that sustain society~\cite{mas2017land}. As a result, timely land cover monitoring and classification are of crucial importance for assessing gradual degradation-ecosystem processes. Furthermore, it is important to be in line with the United Nations Sustainable Development Goals (SDGs) specifically SDG 15 concerning “Life on Land” \cite{tulbure2021regional}.

In recent years, Deep Learning (DL) outpacing the other machine learning techniques remote sensing for the LULC classification~\cite{vali2020deep}. However, there are still big challenges to solve for real-world data, i.e. highly imbalanced and heterogeneous datasets,  big-size areas for classification, low spatial-resolution images, and noisy labels. In addition, for LULC classification it is commonly used multi-channel and multi-spectral images. In this scenario, it is difficult to handle the limited GPU memory or to follow the classical RGB transfer learning approach using pre-trained ConvNet deep models and use limited GPU memory. These limitations reveal huge opportunities to propose new  DL tools.

In this context, we present a real-world data analysis using 13 channels and 17 classes. We have used only open datasets: Multi-seasonal Landsat 8 (cloud-free) from the central west region of Mexico (Jalisco),  in combination with the terrain data and the LULC map 2016 of the same region (with labels). In our opinion, the most important result of the analysis is the identification of four similar classes. In addition, a novel lightweight model is proposed to classify the land use according to specific challenges: multi-class classification, Minimum Mapping Unit (MMU), and small size multi-channel input data. The numerical results in the test set confirm that this strategy achieves a remarkable performance for three of  17 classes and overall accuracy of 0.83 in the coarse classification. 

The rest of the paper is organized as follows. In Section \ref{sec:previous} we collect previous and related works.  The datasets used and the experimental setup are described in sections \ref{sec:data} and \ref{sec:exp_setup}. After, the collecting the results and their analysis are described in Section \ref{sec:results}.  The discussion is presented in Section \ref{sec:discussion}. We devote Section \ref{sec:conlusions} to state our conclusions. Finally, the paper ends with the acknowledgments  and a list of cited references.

\section{Background}

The land-use classes are considered the human-made areas such as roads, agriculture, or cities, while land-cover classes are related to natural earth resources (e.g., water, forests, etc)~\cite{alhassan2019deep}. Remote Sensing (RS) data have been widely used in the last decades to elaborate LULC maps \cite{thenkabail2016remote} \cite{manakos2014land}. One of the main factors that make it difficult to monitor land cover changes is spectral confusion: different types of LULC can present similar spectral responses, and the same type of LULC can present different spectral responses depending on phenology, conservation state, and density. Due to rapid LULCC, frequent updating of maps is needed. Moreover, the elaboration of multi-date cartographic databases is required to assess change \cite{radoux2010automated}.

\begin{figure}
\begin{center}
\includegraphics[width=0.95\textwidth]{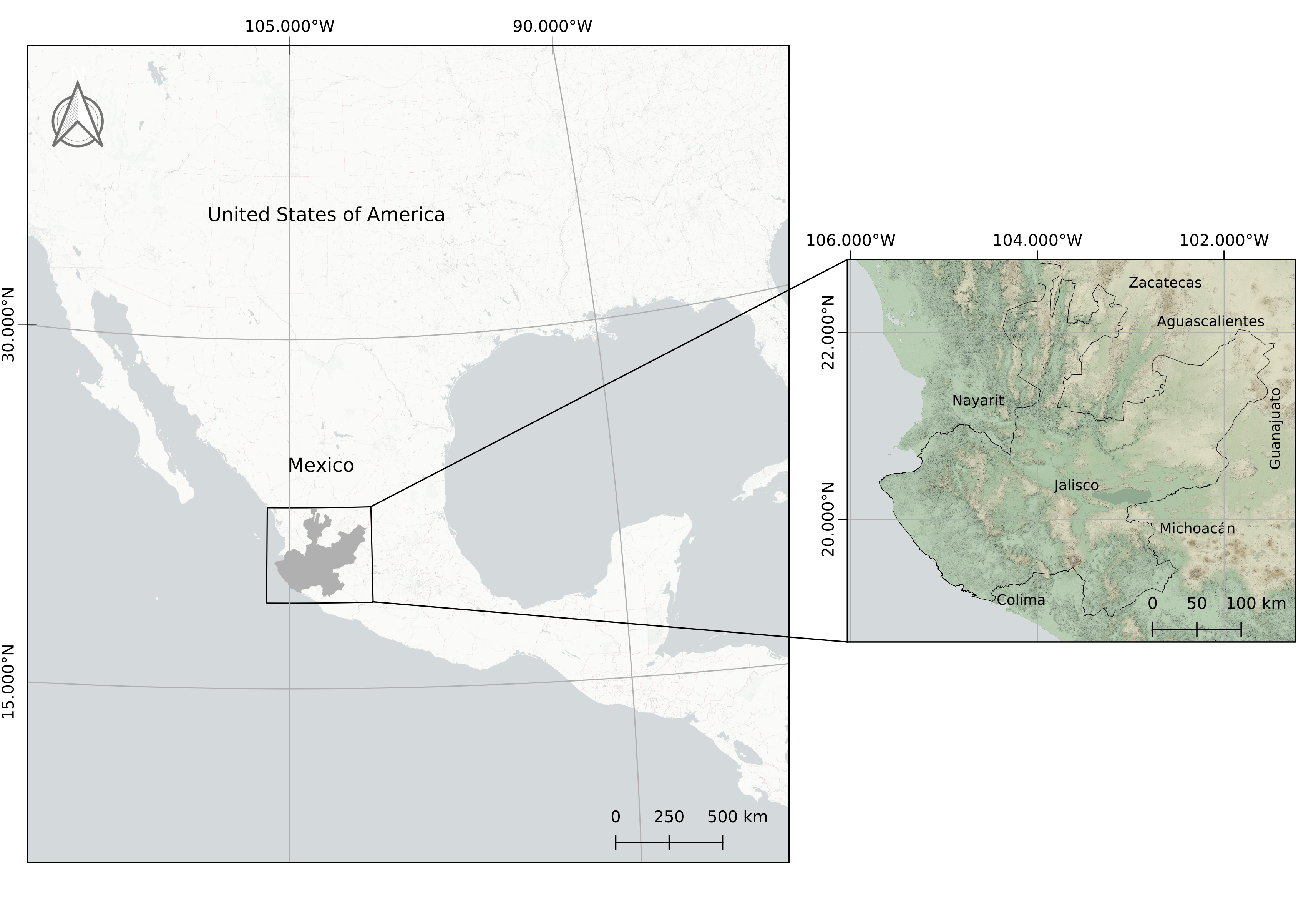}
\end{center}
   \caption{Study area: Jalisco, Mexico.}
\label{fig:Study Area}
\end{figure}

The study area is the state of Jalisco which is located in central-western Mexico between latitudes 22° 45' and 18°55' north, and longitudes 101° 28' and 105° 42' west  (see Figure \ref{fig:Study Area}), with an area of about $80,200{km}^2$. The climates are temperate, tropical rainy and dry \cite{sanchez2013actualizacion}. Altitudes range from 0 to 4,300 meters above sea level. The dominant vegetation in the state is deciduous and subdeciduous forest, The temperate region of the state presents oak, and pine, finally, in the northern and northwestern regions there are areas of natural grasslands. The resulting (complex) environment requires environmental monitoring at a fine scale.

\section{Related works} \label{sec:previous}

The most common use of deep learning in the classification of satellite images is using high-quality images and low number of classes such as images  EUROSAT (10 classes)~\cite{helber2019eurosat} or LandCoverNet (7 classes) \cite{alemohammad2020landcovernet} based on sentinel-2 (10 m spatial resolution). Our work is different from them according to the following aspects: 

\begin{enumerate}
    \item Jalisco data is more diverse, not only with more classes but also, have a wide range of LC and LU over the region. These characteristics have been shown that could affect the class separability~\cite{tulbure2021regional}.
    \item We use low spatial resolution images mainly Lansat8 images with 30 m of spatial resolution, in contrast  to sentinel-2 spatial resolution (10 m). 
    \item Our real-world base map is semiautomatic-labeled using the methodology~\cite{olofsson2014good}  and this requires considerable efforts to obtain a reliable accuracy for the Jalisco region the overall accuracy is  89.48\%.
    \item We use a large number of channels combining satellite bands and terrain data. At the present, most of the DL architectures use three channels (RGB) and usually are pre-trained with IMAGENET dataset. The deep (number of layers)  of those networks are incompatible with the objective of monitoring the MMU.
    \item The multi-seasonal data from Landsat 8 has a coarse temporal resolution due to the cloud cover.   
    \item Usually, most LC datasets were created with big homogeneous generic areas (water vs non-water, forest vs. non-forest). This is because, at global extents, classes are heterogeneous internally, and merging them becomes necessary for achieving reasonable accuracies~\cite{tulbure2021regional,abramovich2019classification}. It is important to remark that these characteristics are not compatible with the heterogeneous distribution of  Jalisco's region an either with LC regional requirements. 
\end{enumerate}

Regarding the classification performance, in our opinion, the fairest comparison of our results is the approach proposed by the MAD-MEX methodology~\cite{gebhardt2014mad}. Due to they used 12 classes and  Landsat 8 images. However, it is important to note that they used a combination of Segmentation for Object-Based Image Analysis (OBIA) and traditional machine learning techniques. The overall accuracy reported from that paper is 76 \%. In addition,  it is important to give context about the typical performance values. According to~\cite{mas2016comment} the accuracy assessment in the MAD-MEX methodology, the tropical evergreen forest was very accurately classified when the accuracy $>$ 75 \%; moderate results were obtained for the tropical deciduous forest when the accuracy is around 70 \%, and temperate forest categories were classified poorly (accuracy indices ranging from 50 \% to 60 \%).

\section{Data}\label{sec:data}

\subsection{LULC map 2016}

We use the land use cover map from Jalisco 2016 \cite{mapa-2016} which is public and can be consulted
\href{https://datos.jalisco.gob.mx/dataset/mapa-coberturas-del-suelo-estado-de-jalisco-al-2016}{here}. This map was developed jointly by the National Forestry Commission of Mexico (CONAFOR) and the Jalisco State Government. The resulting 2016 map is composed of 24 thematic categories besides, accuracy was evaluated following the methodology proposed by  Olofsson~\cite{olofsson2014good}  with an overall accuracy of {89.48 $\pm$ 0.2}{\%}.

\subsection{Landsat 8 cloud-free}

We use  Landsat 8 data from January 1 to December 31, 2016, were collected through Google Earth  Engine (GEE)\cite{gorelick2017google}, the preprocessing script is available  at the following \href{https://code.earthengine.google.com/7f1c99a06218e52c390168d2d3899080 
}{link}. The cloud-free composite was generated by filtering the images by the values of the quality layer, where Normalized Difference Vegetation Index (NDVI) and Normalized Difference Water Index (NDWI) by applying formulas \ref{ndvi} and \ref{ndwi} on the corresponding bands. 
\begin{equation}
    \label{ndvi}
     NDVI = \frac{NIR - RED}{NIR + RED}
\end{equation}

\begin{equation}
    \label{ndwi}
    NDWI  = \frac{NIR - SWIR}{NIR + SWIR}
\end{equation}
where $NIR$, $NDWI$, $SWIR$  and $RED$ are   defined in Table \ref{tab:canales1}. 

\begin{table}
\begin{center}
\begin{tabular}{ll}
\toprule
Channel & Data  \\
\midrule
1 & Band 2 (BLUE) SR  0.452-0.512 $\mu m$ \\
2 & Band 3 (GREEN) SR  0.533-0.590 $\mu m$ \\
3 & Band 4 (RED) SR 0.636-0.673$\mu m$ \\
4 & Band 5 (NIR near infrared)  SR  0.851-0.879 $\mu m$ \\
5 & Band 6 (SWIR Shortwave infrared) SR 0.636-0.673 $\mu m$ \\
6 & Band 7 (SWIR 2) SR 2.107-2.294 $\mu m$ \\
7 & NDVI =  (Band 5 – Band 4) / (Band 5 + Band 4) \\
8 & NDWI =  (Band 6 – Band 5) / (Band 6 + Band 5) \\
\bottomrule
\end{tabular}
\caption{Landsat8 channels. Where  SR = Surface Reflectance. }
\label{tab:canales1}
\end{center}
\end{table}

\subsection{Relief data}

A digital elevation model was included which was obtained through the portal of the National Institute of Statistics and Geography (\href{https://www.inegi.org.mx/app/geo2/elevacionesmex/}{INEGI}) with a spatial resolution of 30 meters, additionally, from this data, we derived the slope, aspect, tangential and profile curvature. The GRASS software \cite{GRASS_GIS_software} was used for this purpose.   

\subsection{Base line dataset}

\begin{figure}[t]
\begin{center}
\includegraphics[width=0.75\textwidth]{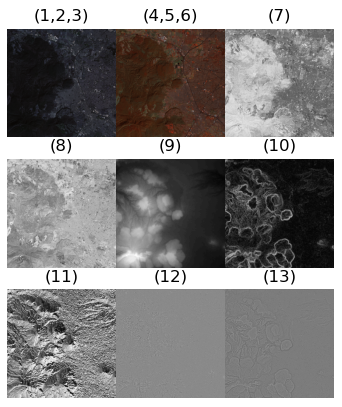}
\end{center}
   \caption{Example of the staked dataset image channels. We enumerate  the number of the channels at the top of each image.}
\label{fig:region}
\end{figure}

Landsat 8,  relief and LULC map 2016 (Labels) data were stacked into a single dataset (13 channels) an example of the data normalized can be seen in Figure \ref{fig:region}. The name of each channel is detailed in Table \ref{tab:canales2}. In addition, the name of each 17 LULC map 2016 class is presented in Table \ref{tab:LULCclasses}, the reduction in the number of class from 24 to 17 is due to the fact that the categories of permanent and perennial agriculture are merged, as is the case for induced and cultivated grassland. Another 5 categories were not considered as they were already well represented in the landscape.

The approach for patch creation was to take only $3\times3\times13$ pixel groups where all pixels belong to a single category, under this approach to achieve a balanced sample with an equal number of elements it was necessary to reduce the number of classes to be analyzed to 17 (see Table \ref{tab:LULCclasses}), this resulted in a total of 21,046 patches with 13 channels, further divided into 70 \% for training, 15 \% for validation and 15 \% for test.

\begin{table}[h]
\begin{center}
\begin{tabular}{ll}
\toprule
Channel & Data  \\
\midrule
1 & Lsat8 Band 2 (BLUE)  \\
2 & Lsat8 Band 3 (GREEN)  \\
3 & Lsat8 Band 4 (RED)  \\
4 & Lsat8 Band 5 (NIR near infrared)   \\
5 & Lsat8 Band 6 (SWIR Shortwave infrared)  \\
6 & Lsat8 Band 7 (SWIR 2)  \\
7 & Lsat8 NDVI =  (Band 5 – Band 4) / (Band 5 + Band 4) \\
8 & Lsat8 NDWI =  (Band 6 – Band 5) / (Band 6 + Band 5) \\
9 & Relief DEM: Digital elevation model   \\
10 & Relief Slope: Slope - Degress \\
11 & Relief Aspect - Degress \\
12 & Relief Tangencial Curvature \\
13 & Relief Profile Curvature \\
\bottomrule
\end{tabular}
\caption{Band or channel used in the Baseline dataset.  Lsat8 = Landsat8. }
\label{tab:canales2}
\end{center}
\end{table}

\begin{table}[h]
\begin{center}
\begin{tabular}{lll}
\toprule
Index & ID & Class \\
\midrule
0& 32 & Water \\
1& 2  & Coniferous forest \\
2& 1  & Upland coniferous forest  \\
3& 3  & Oak forest and riparian forest \\
4& 7  & Cloud forest and low evergreen forest \\
5& 9  & Mangrove and peten \\
6& 15 & Crassicaule schrub \\
7& 5  & Mezquital and submontane shrub \\
8& 34 &  Cultivated and induced grasslands  \\
9& 28 & Natural grasslands   \\
10& 12 &  Tropical dry forest \\
11& 13 &  Tropical semideciduous forest  \\
12& 31 & Bare land \\
13& 29 & Rain fed agriculture \\
14& 35 & Cropland irrigated \\
15& 30 & Urban areas \\
16& 26 & Hydrophilic halophilic vegetation \\
\bottomrule
\end{tabular}
\end{center}
\caption{Land cover and land use classes used as base-line.}
\label{tab:LULCclasses}
\end{table}

\section{LUCC Lightweight ConvNet}
\begin{figure}
\begin{center}
\includegraphics[width=0.25\textwidth]{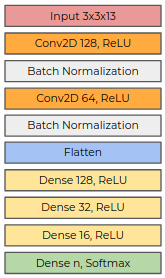}
\end{center}
   \caption{The proposed lightweight LUC ConvNet architecture. }
\label{fig:cnn}
\end{figure}

The proposed ConvNet architectures and their hyperparameters were designed  taking into account, the following:
\begin{enumerate}
    \item Classify  the  LC using MMU.
    \item The possibility to handle 2D signal with 13 channels 
    \item Light weights  to reduce the inference and training time.
\end{enumerate}
To solve these requirements we combine our experience with an expert-knowledge, tests and  image analysis to select the most appropriate architecture. As a result,  we propose ConvNet architecture  presented in Figure \ref{fig:cnn}. In this novel ConvNet we propose an input shape of $3\times 3\times 13$. In consequence, it is possible to classify the MMU. Following, we use two convolution layers with 128 and 64 filters (a wide architecture), with window size of $1\times1$ and ReLU as activation function. The $1\times1$ convolution is used as channel-wise pooling to promote promote learning across channels such as \cite{lin2013network,szegedy2014going,he2015deep} and help us to  handle and learn using the 13 channels. After the ConvNet layers, a Batch-normalization layer have added~\cite{DBLP:journals/corr/IoffeS15} to reduce the \textit{internal covariate shift} problem, in fact,  we expect to have this problem due to the study area having small-distributed examples unevenly across its territory. Then, following the \textit{flatten} layer,  we  decide to add three dense (fully connected) layer  [128,32,16] with ReLU, each one have a Gaussian dropout (30\%) to reduce the overfitting~\cite{labach2019survey}.

\section{Experimental setup}\label{sec:exp_setup}

We made three trainings with the proposed architecture: i) a base-line train using a dataset with 17 balanced-classes, presented at Table~\ref{tab:LULCclasses}, ii) coarse-grain training, due to  embedded analysis, bring to light that eighth classes could be grouped into four, reducing to 13 classes  iii) fine-grain training  for binary classification in each group (with similar classes). The hyperparameters values in all the training are: learning rate= 0.0001, epochs= 150  and  batch size= 32. In addition, we used  0-$90^{o}$ random rotations,  horizontal and vertical flips as data augmentation.

\begin{figure}
\begin{center}
\includegraphics[width=0.25\textwidth]{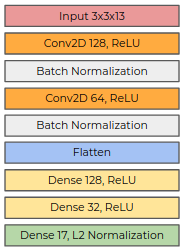}
\end{center}
   \caption{The proposed lightweight LUC ConvNet architecture form embedded  analysis. }
\label{fig:cnn2a}
\end{figure}

The embedded analysis  is based on a \textit{embedded layer} replacing the last two layers in our CNN architecture  for a embedded layer consists of an output dimension of 17 and an L2 normalization (See Figure ~\ref{fig:cnn2a}).  Once the model is trained, we apply  a t-distributed stochastic neighbor embedding  (t-SNE)~\cite{hinton2002stochastic} on the latent ConvNet representation. 

\section{Results and analysis}\label{sec:results}
The base-line test-loss  and  test-accuracy  (using 17 classes) are 0.81 and 0.73 respectively (using the test data split). The details of the base-line classification performance per class are presented in Table~\ref{tab:report1}.  We observe that class-ID 32 (water) has the best classification performance with 0.96 of F1-score. In contrast, the id-classes 2,3,12,15,28,29,34 and 35 has the worst classification values in range of f1-score values from 0.35 to 0.75. In addition, the confusion matrix presented at Table~\ref{tab:conf_matrix1} help us to detect the most frequent error class.

\begin{figure}
\begin{center}
\includegraphics[width=0.75\textwidth]{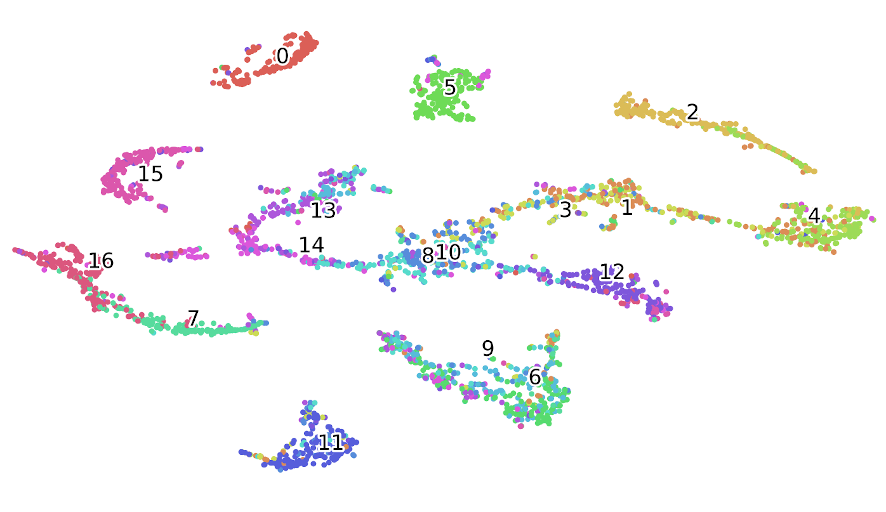}
\end{center}
   \caption{Embedding representation of the base-line training. The closer groups are  g1=[(index = 1: class ID =2 Coniferus forest)(index 3 :  ID 3 Oak forest)], g2 = [(index  8 :  ID 34 cultivated grassland) (index 10 :  ID 12 Tropical dry forest)] , g3=[(index 13: ID 29 rain feed agriculture)(index 14 :  ID 35 cropland irrigated)], g4=[(index 6, ID 15 crassicaule scrub) (index 9: ID 28 natural grassland)].}
\label{fig:emb}
\end{figure}

The proposed analysis using the embedding  representation   is presented at Figure~\ref{fig:emb}. By visualizing in 2D, could be see how useful are the representations learned by ConvNet
to distinguish between the classes in familiar and new domains. In this 2D representation, we can observe the presence of category clusters formed by the classes [(index = 1: class ID =2 Coniferus forest),(index 3 :  ID 3 Oak forest)], [(index  8 :  ID 34 cultivated grassland), (index 10 : ID 12 Tropical dry forest)] ,  [(index 13: ID 29 rain feed agriculture),(index 14 :  ID 35 cropland irrigated)], [(index 6, ID 15 crassicaule scrub) (index 9: ID 28 natural grassland)]. As a result, we decide to create new four groups: g1=(ID 2, ID 3), g2=(ID 34, ID12), g3=( ID 29, ID 35), and g4=(ID 15, ID 28). 



\begin{table}
\begin{center}
\setlength\tabcolsep{2pt}
\begin{tabular}{cccccl} 
\toprule
Precision & Recall & F1-score &  Support  &  ID & Class \\
\midrule
 0.98 & 0.93 & 0.96 & 207 & 32 & Water \\
 0.60 & 0.46 & 0.52 & 185 & 2  & Coniferous forest \\
 0.89 & 0.95 & 0.92 & 210 & 1  & Upland coniferous forest \\
 0.42 & 0.30 & 0.35 & 180 & 3  & Oak forest and riparian forest  \\
 0.69 & 0.81 & 0.75 & 196 & 7  & Cloud forest and low evergreen forest \\
 0.93 & 0.99 & 0.96 & 169 & 9  & Mangrove and peten \\
 0.60 & 0.85 & 0.71 & 183 & 15 & Crassicaule schrub  \\
 0.89 & 0.76 & 0.82 & 192 & 5  & Mezquital and submontane shrub \\
 0.45 & 0.48 & 0.46 & 168 & 34 & Cultivated and induced grasslands \\
 0.43 & 0.36 & 0.39 & 185 & 28 & Natural grasslands  \\
 0.59 & 0.58 & 0.58 & 183 & 12 & Tropical dry forest \\
 0.74 & 0.94 & 0.83 & 206 & 13 & Tropical semideciduous forest \\
 0.74 & 0.69 & 0.72 & 183 & 31 & Bare land \\
 0.60 & 0.47 & 0.52 & 191 & 29 & Rain fed agriculture \\
 0.76 & 0.74 & 0.75 & 170 & 35 & Cropland irrigated \\
 0.79 & 0.84 & 0.81 & 177 & 30 & Urban areas \\
 0.80 & 0.88 & 0.84 & 172 & 26 & Hydrophilic halophilic vegetation \\
\bottomrule
\end{tabular}
\end{center}
\caption{Base-line classification report over the 17 classes.}
\label{tab:report1}
\end{table}

\begin{table}
\begin{center}
\setlength\tabcolsep{1.1pt}
\begin{tabular}{ccccccccccccccccc|c}
\toprule
\small
 32 & 2 & 1 & 3 & 7 & 9 & 15 & 5 & 34 & 28 & 12 & 13 & 31 & 29 & 35 & 30 & 26 & ID \\
\midrule
192 & 0 & 0 & 0 & 0 & 1 & 0 & 0 & 1 & 1 & 6 & 0 & 1 & 3 & 0 & 1 & 1 & 32 \\
0 & 85 & 6 & 29 & 36 & 0 & 3 & 0 & 4 & 3 & 10 & 7 & 0 & 1 & 1 & 0 & 0 & 2 \\
0 & 0 & 200 & 0 & 10 & 0 & 0 & 0 & 0 & 0 & 0 & 0 & 0 & 0 & 0 & 0 & 0 & 1 \\
0 & 28 & 0 & 54 & 15 & 0 & 4 & 0 & 31 & 15 & 12 & 16 & 2 & 3 & 0 & 0 & 0 & 3 \\
0 & 8 & 14 & 4 & 158 & 0 & 0 & 0 & 0 & 0 & 0 & 7 & 1 & 0 & 4 & 0 & 0 & 7 \\
0 & 0 & 0 & 0 & 0 & 167 & 0 & 0 & 0 & 0 & 2 & 0 & 0 & 0 & 0 & 0 & 0 & 9 \\
0 & 1 & 0 & 2 & 0 & 0 & 156 & 0 & 0 & 16 & 1 & 0 & 1 & 6 & 0 & 0 & 0 & 15 \\
0 & 2 & 0 & 0 & 0 & 0 & 16 & 145 & 0 & 1 & 3 & 0 & 0 & 0 & 4 & 0 & 21 & 5 \\
0 & 3 & 0 & 9 & 1 & 1 & 6 & 0 & 80 & 25 & 15 & 9 & 10 & 8 & 1 & 0 & 0 & 34 \\
0 & 4 & 4 & 13 & 0 & 0 & 45 & 0 & 19 & 67 & 3 & 0 & 2 & 23 & 3 & 2 & 0 & 28 \\
0 & 5 & 0 & 12 & 1 & 4 & 6 & 1 & 16 & 4 & 106 & 21 & 4 & 1 & 2 & 0 & 0 & 12 \\
0 & 0 & 0 & 3 & 3 & 1 & 0 & 0 & 0 & 0 & 5 & 194 & 0 & 0 & 0 & 0 & 0 & 13 \\
1 & 1 & 0 & 2 & 0 & 0 & 0 & 1 & 6 & 6 & 6 & 2 & 127 & 3 & 1 & 14 & 13 & 31 \\
2 & 2 & 0 & 0 & 0 & 0 & 13 & 3 & 18 & 14 & 4 & 4 & 4 & 89 & 20 & 17 & 1 & 29 \\
0 & 2 & 0 & 1 & 4 & 2 & 7 & 0 & 4 & 2 & 6 & 1 & 4 & 6 & 125 & 5 & 1 & 35 \\
0 & 0 & 0 & 0 & 0 & 1 & 3 & 2 & 0 & 2 & 2 & 0 & 11 & 5 & 3 & 148 & 0 & 30 \\
0 & 0 & 0 & 0 & 0 & 2 & 0 & 11 & 0 & 0 & 0 & 0 & 4 & 1 & 1 & 1 & 152 & 26 \\
\bottomrule
\end{tabular}
\end{center}
\caption{Base-line confusion matrix, using 17 classes.}
\label{tab:conf_matrix1}
\end{table}

The results of the coarse-grain classification with 13 classes (using the same ConvNet) shows lower test-loss value, 0.53, and higher test-accuracy, 0.83 in comparison base-line classification (with the 17 classes). The  coarse-grain classification details are presented at Table \ref{tab:report_groups}. Note that the four groups increase their performance substantially. As expected, the coarse-grain confusion matrix presented at Table \ref{tab:conf_matrix_groups} shows noteworthy differences in the diagonal. As a result, the  same model in coarse-grain classification has fewer classification mistakes.

\begin{table}
\begin{center}
\setlength\tabcolsep{2.5pt}
\begin{tabular}{cccccl} 
\toprule
Precision &   Recall & F1-score &  Support  &  ID & Class \\
\midrule
0.97 & 0.97 & 0.97 & 152 & 32 & Water \\
0.83 & 0.98 & 0.90 & 199 & 1 & Upland coniferous forest \\
0.85 & 0.81 & 0.83 & 192 & 7 & Cloud forest and low evergreen forest  \\
0.74 & 0.70 & 0.72 & 184 & g1 & Coniferus forest and  Oak forest \\
0.64 & 0.52 & 0.57 & 170 & g2 &  Cultivated grassland  and Tropical dry forest \\
0.83 & 0.62 & 0.71 & 186 & g3 & Rain feed agriculture and Cropland irrigated \\
0.70 & 0.86 & 0.77 & 184 & g4 & Crassicaule scrub and  Natural grassland \\
0.97 & 1.00 & 0.98 & 190 & 9 & Mangrove and peten \\
0.87 & 0.87 & 0.87 & 193 & 5 & Mezquital and submontane shrub \\
0.83 & 0.95 & 0.88 & 202 & 13 & Tropical semideciduous forest \\
0.82 & 0.73 & 0.77 & 173 & 31 & Bare land \\
0.83 & 0.87 & 0.85 & 190 & 30 & Urban areas \\
0.89 & 0.88 & 0.88 & 200 & 26 & Hydrophilic halophilic vegetation  \\
\bottomrule
\end{tabular}
\end{center}
\caption{Classification report of coarse-grain  (with grouped classes).}
\label{tab:report_groups}
\end{table}

\begin{table}
\begin{center}
\setlength\tabcolsep{1pt}
\begin{tabular}{ccccccccccccc|c}
\toprule
\small
32 & 1 & 7 & g1 & g2 & g3 & g4 & 9 & 5 & 13 & 31 & 30 & 26 & ID \\
\midrule
148 & 0 & 0 & 0 & 0 & 1 & 0 & 0 & 0 & 0 & 2 & 0 & 1 & 32 \\
0 & 196 & 3 & 0 & 0 & 0 & 0 & 0 & 0 & 0 & 0 & 0 & 0 & 1 \\
0 & 30 & 155 & 4 & 0 & 0 & 0 & 0 & 0 & 3 & 0 & 0 & 0 & 7 \\
0 & 7 & 17 & 128 & 12 & 3 & 8 & 0 & 2 & 6 & 0 & 1 & 0 & g1 \\
0 & 0 & 3 & 30 & 88 & 2 & 6 & 2 & 2 & 27 & 9 & 1 & 0 & g2 \\
0 & 0 & 2 & 2 & 7 & 115 & 42 & 0 & 2 & 0 & 0 & 13 & 3 & g3 \\
0 & 4 & 0 & 6 & 6 & 6 & 158 & 0 & 1 & 0 & 0 & 3 & 0 & g4 \\
0 & 0 & 0 & 0 & 0 & 0 & 0 & 190 & 0 & 0 & 0 & 0 & 0 & 9 \\
0 & 0 & 0 & 0 & 2 & 1 & 11 & 0 & 168 & 0 & 0 & 0 & 11 & 5 \\
0 & 0 & 2 & 0 & 7 & 0 & 0 & 2 & 0 & 191 & 0 & 0 & 0 & 13 \\
2 & 0 & 0 & 1 & 14 & 3 & 2 & 0 & 0 & 4 & 126 & 15 & 6 & 31 \\
0 & 0 & 0 & 0 & 1 & 6 & 0 & 1 & 1 & 0 & 14 & 166 & 1 & 30 \\
2 & 0 & 0 & 1 & 1 & 1 & 0 & 1 & 17 & 0 & 2 & 0 & 175 & 26 \\
\bottomrule
\end{tabular}
\end{center}
\caption{Coarse-grain confusion matrix report.}
\label{tab:conf_matrix_groups}
\end{table}

The corresponding test-loss and test-accuracy are: g1 loss= 0.55, acc= 0.70, g2 loss= 0.38, acc= 0.85, g3 loss= 0.35, acc= 0.86 and g4 loss= 0.45, acc= 0.80.  The fine-grain  classification details over the four groups are presented in Table \ref{tab:report_group1}. We observe from this table that the g1 group (2: Coniferous forest, 3:Oak forest) has the lowest classification performance. In Table~\ref{tab:conf_matrix_group1} we present the confusion matrices  of each group. The values are barely in line with the classification report.

\begin{table}
\begin{center}
\setlength\tabcolsep{2pt}
\begin{tabular}{cccccl} 
\toprule
Precision &   Recall & F1-score &  Support  &  ID & Class \\
\midrule
0.72 & 0.64 & 0.68 & 184 & 2 & Coniferous forest \\
0.68 & 0.76 & 0.72 & 188 & 3 & Oak forest \\
\midrule
0.83 & 0.87 & 0.85 & 190 & 34 & Cultivated grassland \\
0.86 & 0.82 & 0.84 & 182 & 12 & Tropical dry forest \\
\midrule
0.88 & 0.82 & 0.85 & 182 & 29 & Rain feed agriculture  \\
0.84 & 0.89 & 0.86 & 190 & 35 & Cropland irrigated \\
\midrule
0.84 & 0.75 & 0.79 & 194 & 15 & Crassicaule scrub  \\
0.76 & 0.84 & 0.80 & 178 & 28 & Natural grassland \\
\bottomrule
\end{tabular}
\end{center}
\caption{Fine-grain classification report per group.}
\label{tab:report_group1}
\end{table}

\begin{table}
\begin{center}
\setlength\tabcolsep{1pt}
\begin{tabular}{ccc|ccc|ccc|ccc}
\toprule
2 & 3 & ID& 34 & 12 & ID & 29 & 35 & ID & 15 & 28 & ID\\
\midrule
118 & 66 & 2 & 166 & 24 & 34 & 150 & 32 & 29 & 146 & 48 & 15 \\
46 & 142 & 3 & 33 & 149 & 12 & 21 & 169 & 35 & 28 & 150 & 28 \\
\bottomrule
\end{tabular}
\end{center}
\caption{Fine-grain confusion matrix.}
\label{tab:conf_matrix_group1}
\end{table}

\begin{figure}
\begin{center}
\includegraphics[width=0.43\textwidth]{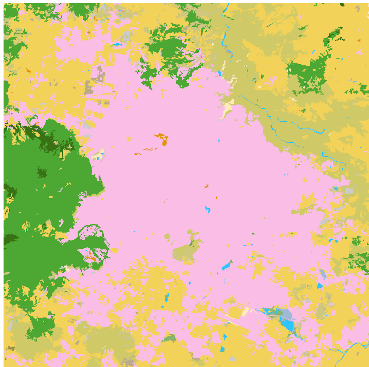}
\includegraphics[width=0.43\textwidth]{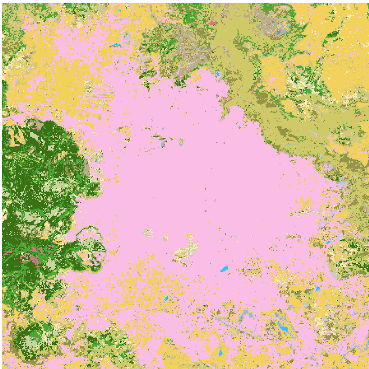}
\end{center}
   \caption{Left: Ground truth map of the urban area of Guadalajara. Right:  prediction over the same area. }
\label{fig:map_pred}
\end{figure}

In Figure \ref{fig:map_pred}  is presented a ground truth map and the prediction in the Guadalajara urban area using the coarse-grain data model. It is possible to note that the urban area is relatively well classified, in contrast,  this model had a lower performance to discriminate between the land cover classes. It is possible to see some \textit{salt and paper} artifacts in the prediction, this artifacts are related with the mask size mainly.


\section{Discussion}\label{sec:discussion}

Generally speaking, the results and analysis presented in the previous section  indicate that the classification performance are competitive taking into account the context: number of  classes (the number of classes usually have a stronger impact in the classification performance as the number of classes increases \cite{abramovich2019classification}), low spatial resolution (30 m), multiseasonal data, noisy labels (89\% of accuracy). Regarding to classification performance, our overall classification with 13 (and 17) classes outperforms the 75\%  and according to to~\cite{mas2016comment} this could be considered as very accurate result. In addition, the reported accuracy is achieved without any additional post-processing by experts, which is still a common practice in LULC map making a slow and costly process.

The importance of the embedded analysis is not only that this help to increase the classification performance (if the similar classes are grouped), but also, reveal the  most difficult classes where the human photo-interpretation could be more critical.

About the fine-grain classification we find that our worst classification result with g1: Coniferus forest and Oak forest, there other works which have  been reported previosly  confusion between these two classes



\section{Conclusions and Future work} \label{sec:conlusions}

In this paper we have presented a novel lightweight ConVnet to classify the LULC images combining real-world open data sources: Lansat8 (multi-seasonal  and low spatial resolution) cloud-free, realief, and  2016 Jalisco's LULC map. In contrast, to  most common  works,  the proposed ConvNet can handle 13 channels due to $1\times1$ convulution take into account the  relation over the  channels allowing to define $3\times3$ a  minimum mapeable unit of landsat8 images which is useful in the context of landcover clasification.
After an embedded analysis, of the base-line results, we group the similar classes  increase the test-accuracy performance from 0.73 to 0.83. The open challenges of the proposed methos are the salt and paper noise in the prediction and the lower classification performance with the groped classes.

In our opinion, these results represent an excellent initial step toward automatic LUC image classification. Our research possibly supports the decision-makers providing evidence, for example, of deforestation or forest degradation. In our opinion, the main value of our work lies in the fact that we obtain good classification performance using real-world data sources. We hope that this method could be useful by other regions and groups with limited data sources or computational resources. It is important to remark that our investigations into this area are still in progress and we hope to focus on exploring the combination with other deep learning methods such as \cite{moya2021trainable,khan-nui-2021} to increase the performance and make robust  classification. 






\section*{Acknowledgments}
The authors would like to thank to SEMADET and CONAFOR. 
\bibliographystyle{unsrt}  
\bibliography{references}

\end{document}